\documentclass[10pt, a4paper]{article}
\usepackage{lrec}
\usepackage{multibib}
\newcites{languageresource}{Language Resources}
\usepackage{graphicx}
\usepackage{amssymb,amstext,amsmath,amsthm}
\usepackage{tabularx}
\usepackage{booktabs}

\usepackage{epstopdf}
\usepackage[latin1]{inputenc}

\usepackage{hyperref}
\usepackage{xstring}

\newenvironment{enumerate2}
    {\begin{enumerate}
        \vspace{-0.5em}
        \setlength{\abovedisplayskip}{0pt}
        \setlength{\belowdisplayskip}{0pt}
        \setlength{\itemsep}{3pt}
        \setlength{\parskip}{0pt}
        \setlength{\parsep}{0pt}
        \setlength{\topsep}{0pt}
        \setlength{\partopsep}{0pt}
    }
    {\vspace{-0.5em}
    \end{enumerate}}

\title{\textbf{Building a Web-Scale Dependency-Parsed Corpus from Common Crawl}}

\name{Alexander Panchenko$^\ddag$, Eugen Ruppert$^\ddag$, Stefano Faralli$^\dag$, Simone P. Ponzetto$^\dag$, Chris Biemann$^\ddag$}

\address{$\ddag$ University of Hamburg, Department of Informatics, Language Technology Group, Germany \\ $\dag$ University of Mannheim,  School of Business Informatics and Mathematics, Data and Web Science Group, Germany \\
\url{{panchenko,ruppert,biemann}@informatik.uni-hamburg.de} \\
\url{{stefano,simone}@informatik.uni-mannheim.de} \\
}

\abstract{We present \textsc{DepCC}, the largest-to-date linguistically analyzed corpus in English including 365 million documents, composed of 252 billion tokens and 7.5 billion of named entity occurrences in 14.3 billion sentences from a web-scale crawl of the \textsc{Common Crawl} project. The sentences are processed with a dependency parser and with a named entity tagger and contain provenance information, enabling various applications ranging from training syntax-based word embeddings to open information extraction and question answering. We built an index of all sentences and their linguistic meta-data enabling quick search across the corpus. We demonstrate the utility of this corpus on the verb similarity task by showing that a distributional model trained on our corpus yields better results than models trained on smaller corpora, like Wikipedia. This distributional model outperforms the state of art models of verb similarity trained on smaller corpora on the SimVerb3500 dataset. \\ \newline }

\begin{document}

\maketitleabstract

\section{Introduction}

Large corpora are essential for the modern data-driven approaches to natural language processing (NLP), especially for unsupervised methods, such as word embeddings~\cite{Mikolov:13} or open information extraction~\cite{banko2007open} due to the ``unreasonable effectiveness of big data''~\cite{halevy2009unreasonable}. However, the size of commonly used text collections in the NLP community, such as BNC\footnote{\url{http://www.natcorp.ox.ac.uk}} or Wikipedia is in the range 0.1--3 billion tokens, which potentially limits coverage and performance of the developed models. To overcome this limitation, larger corpora can be composed of books, e.g. in \cite{goldberg2013dataset} a dataset of syntactic $n$-grams\footnote{\url{http://commondatastorage.googleapis.com/books/syntactic-ngrams/index.html}} was built from the 345 billion token corpus of the Google Books project.\footnote{\url{https://books.google.com}} However, access to books is often restricted, which limits use-cases of book-derived datasets. Another source of large amounts of texts is the Web. Multiple researchers investigated the use of the Web for construction of text corpora, producing resources, such as \textsc{PukWaC}~\cite{baroni2009wacky} (2 billion of tokens) and \textsc{ENCOW16}~\cite{Schaefer2015b}  (17 billion of tokens), yet the size of these corpora is still at least one order of magnitude smaller than the web-scale crawls, e.g. \textsc{ClueWeb}\footnote{\url{http://lemurproject.org/clueweb12}} and \textsc{Common Crawl}\footnote{\url{http://www.commoncrawl.org}}. On the other hand, directly using the web crawl dumps is problematic for  researchers as: (1) the documents are not preprocessed, containing irrelevant information, e.g. HTML markup; (2) big data infrastructure and skills are required; (3) (near)duplicates of pages disbalance the corpus; (4) documents are not linguistically analyzed, thus only shallow models can be used. The mentioned factors substantially limit the use of web-scale corpora in natural language processing research and applications. 

The objective of this work is to address these issues and \textit{make access to web-scale corpora a commodity} by providing a web-scale corpus that is ready for NLP experiments as it is linguistically analyzed and cleansed from noisy irrelevant content. Namely, in this paper, we present a technology for constructing of linguistically analyzed corpora from the Web and release \textsc{DepCC}, the largest-to-date dependency-parsed corpus of English texts.

The \textsc{Common Crawl} project regularly produces web-scale crawls featuring a substantial fraction of all public web pages. For instance, as of October 2017, the estimated number of pages on the Web is 47 billion\footnote{\url{http://www.worldwidewebsize.com} at 02.10.2017}, while the corresponding crawl contains over 3 billion pages. To put this number into perspective, according to the same source, the indexed Web contains about 5 billion pages. 

\textsc{Common Crawl} provides the data in the Web ARChive (WARC)  format. Each crawl is provided in the raw form features full HTML pages with metadata or in the form of preprocessed  archives containing texts (WET). The WET archives contain extracted plaintext from the raw crawls. For instance, the 29.5 Tb raw crawl archive (cf. Table~\ref{tab:stages}) has a corresponding 4.8 Tb WET  version with texts. The preprocessing used for producing the WET archives is limited to removal of HTML tags. After a manual check, we also noticed that in WET archives (1) some documents still contain HTML markup; (2) the archives contain document duplicates;  (3) documents are written in various languages making it difficult to train language-specific linguistic models. Finally, most importantly, the WET dumps are not linguistically analyzed, which significantly limits their utility for language processing applications.

\begin{table*}
\footnotesize
\centering
\begin{tabular}{lllllllll}
\toprule

& WaCkypedia & Wikipedia & PukWaC & GigaWord &  ENCOW16 & ClueWeb12 &  Syn.Ngrams & \textsc{DepCC}  \\ \midrule
Tokens (billions) &  0.80 & 2.90 & 1.91 & 1.76 & 16.82 & N/A & 345.00 & 251.92  \\
Documents (millions) & 1.10 & 5.47 & 5.69    & 4.11 & 9.22 & 733.02 & 3.50 & 364.80 \\ 
Type & Encyclop. & Encyclop. & Web & News & Web & Web & Books & Web  \\ 
Source texts & Yes & Yes &Yes &Yes &Yes &Yes & No &Yes \\ 
Preprocessing & Yes & No & Yes & No & Yes & No & No & Yes  \\ 
NER & No & No & No & No & Yes & No & No & Yes  \\ 
Dependency-parsed & Yes & No & Yes & No & Yes & No & Yes & Yes \\

\bottomrule
\end{tabular}
\caption{Comparison of existing large text corpora for English with the \textsc{DepCC} corpus.}
\label{tab:stats}
\end{table*}

In this work, we address these limitations by constructing a text corpus from \textsc{Common Crawl}, which is filtered from irrelevant and duplicate documents and is linguistically analyzed. Namely, the contributions of this paper are the following:

\begin{enumerate2}
\item We present a \textit{methodology} for the creation of the text corpus from the web-scale crawls of \textsc{Common Crawl}. 
\item We present a \textit{software} implementing the methodology in a scalable way using the MapReduce framework.
\item We present the largest-to-date \textit{dependency parsed corpus} of English texts obtained using the developed methodology, also featuring  \textit{named entity tags}. 
\item We show the utility of the web-scale corpora on the \textit{verb similarity} task by outperforming the state of the art on the SimVerb3500 dataset~\cite{gerz-EtAl:2016:EMNLP2016}.
\end{enumerate2}
The corpus and the software tools are available online.\footnote{\url{https://www.inf.uni-hamburg.de/en/inst/ab/lt/resources/data/depcc.html}} Namely, the corpus can be directly used without the need to download it on the Amazon S3 distributed file system, cf. Section~\ref{sec:amazon}\footnote{\url{https://commoncrawl.s3.amazonaws.com/contrib/depcc/CC-MAIN-2016-07/index.html}} The software tools used to build the corpus are distributed under an open source license. The terms of use of the corpus are described in Section~\ref{sec:license}

\section{Related Work}
\label{sec:rel}

\subsection{Large Scale Text Collections}

In Table~\ref{tab:stats}  we compare the \textsc{DepCC} corpus to seven existing large-scale English corpora, described below.

\textsc{WaCkypedia}~\cite{baroni2009wacky} is a parsed version of English Wikipedia as of 2009. The articles are part-of-speech tagged with the TreeTagger~\cite{schmid2013probabilistic} and dependency
parsed with the Malt parser~\cite{nivre2007maltparser}. Similarly to our corpus, the results are presented in the CoNLL format.\footnote{\url{http://www.universaldependencies.org/format.html}} The 2017 version of \textsc{Wikipedia} contains three times more tokens, compared to the version of 2009, yet the there are no distributions of linguistically analyzed recent dumps. \textsc{PukWaC} is a dependency-parsed version of the \textsc{ukWaC} corpus~\cite{baroni2009wacky}, which is processed in the same way as the \textsc{WaCkypedia} corpus.

\textsc{GigaWord}~\cite{parker2011english} is a large corpus of newswire, which is not dependency parsed. The \textsc{ClueWeb12} is a corpus similar to the raw \textsc{Common Crawl} corpus: it contains archives of linguistically unprocessed web pages.

The authors of the \textsc{Google Syntactic Ngrams} corpus~\cite{goldberg2013dataset} parsed a huge collection of books and released a dataset of syntactic dependencies. However, the source texts are not shared due to copyright restrictions, which limits potential use-cases of this resource.

Finally, \textsc{ENCOW16}~\cite{Schaefer2015b} is a large-scale web corpus, which is arguably the most similar one to \textsc{DepCC}. The authors also rely on the Malt parser and perform named entity tagging. However, this corpus contains roughly 15 times less tokens than \textsc{DepCC}.

\begin{table}
\footnotesize
\centering
\begin{tabular}{lr}
\toprule

\textbf{Stage of the Processing} & \textbf{Size (.gz)}  \\ \midrule
Input raw web crawl (HTML, WARC) & 29,539.4 Gb  \\ 
Preprocessed corpus (simple HTML) & 832.0 Gb  \\ 
Preprocessed corpus English (simple HTML) & 683.4 Gb  \\ 
Dependency-parsed English corpus (CoNLL) & 2,624.6 Gb  \\ 
\bottomrule
\end{tabular}
\caption{Various stages of development of the corpus based on the \textsc{Common Crawl} 2016-07 web crawl dump.}
\label{tab:stages}
\end{table}

\begin{figure*}[ht]
    \begin{center}
    \includegraphics[width=1.0\textwidth]{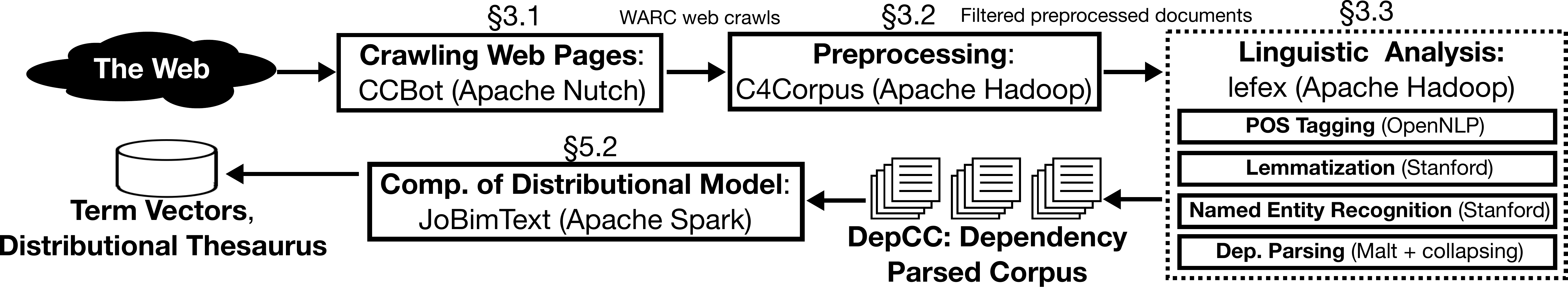}
    \end{center}
    \caption{Outline of the corpus construction approach and experiments described in the paper.}
    \label{fig:outline}
\end{figure*}

\subsection{Common Crawl as a Corpus}
\newcite{kolias2014exploratory} present an exploratory study of one of the early versions of the \textsc{Common Crawl}. The authors provide various descriptive statistics of the dataset regarding language distribution, formats of the documents, etc. 

\textsc{Common Crawl} was used construct a large-scale Finnish  Parsebank consisting of 1.5 billion tokens in 116 million sentences \cite{laippala2014syntactic}. The texts were morphologically and syntactically analyzed. In addition, distributional vector space representations of the
words were obtained using the word2vec toolkit~\cite{Mikolov:13}. The resources were made available under an open license.

GloVe~\cite{pennington2014glove} is an unsupervised model for learning distributional word representations similar to word2vec. The authors distribute\footnote{\url{https://nlp.stanford.edu/projects/glove}} two models trained on the English part of a \textsc{Common Crawl} corpus (comprising respectively 42 and 820 billion of tokens), which are often used to build neural NLP systems, such as~\cite{tsuboi:2014:EMNLP2014}. The models were trained on the \textsc{Common Crawl} documents texts tokenized with the Stanford tokenizer. In addition, the smaller training corpus was lowercased.  

\section{Building a Web-Scale Dependency-Parsed Corpus in English from Common Crawl}

Figure~\ref{fig:outline} shows how a linguistically analyzed corpus is built from the Web. First, web pages are downloaded by the web crawler of \textsc{Common Crawl}, called CCBot. Second, preprocessing, involving elimination of duplicates and language detection, is performed using the C4Corpus tool. Finally, we perform linguistic analysis of the corpus and save the results in the CoNLL format (cf. Section~\ref{sec:format}).

\subsection{Input Web Crawl: the Common Crawl}

The \textsc{DepCC} corpus is based on the crawl of February 2016\footnote{\url{http://commoncrawl.org/2016/02/}} containing more than 1.73 billion URLs. The \textsc{Common Crawl} URL index for this crawl is available online\footnote{\url{http://index.commoncrawl.org/CC-MAIN-2016-07}}, while 
the original files are located in the ``commoncrawl'' bucket on the S3 distributed file system.\footnote{\url{s3://commoncrawl/crawl-data/CC-MAIN-2016-07}} As summarized in Table~\ref{tab:stages}, the total size of the  compressed HTML WARC files is about 30 Tb.

\subsection{Preprocessing of Texts: the C4Corpus Tool}

The raw corpus was processed with the C4Corpus tool~\cite{habernal2016c4corpus} and is available on the distributed cloud-based file system Amazon S3.\footnote{\url{s3://commoncrawl/contrib/c4corpus/CC-MAIN-2016-07}} The tool performs preprocessing of the raw corpus, in five phases:

\begin{enumerate2}
\item Language detection, license detection, and  removal of boilerplate page elements, such as menus;
\item ``Exact match" document de-duplication;
\item Detecting near duplicate documents;
\item Removing near duplicate documents;
\item Grouping the final corpus by language and license.
\end{enumerate2}

The resulting output is a gzip-compressed corpus with a total size of 0.83 Tb (cf. Table~\ref{tab:stages}). For further processing, we selected only English texts with the total size of 0.68 Tb, based on the language detection in the first phase.
Note that we use all texts written in English, not only those published under the CC-BY license.

\begin{table*}[ht]
\footnotesize
\centering
\begin{tabular}{llllllllll}
\toprule

\textbf{ID} & \textbf{FORM} & \textbf{LEMMA} & \textbf{UPOSTAG} & \textbf{XPOSTAG} & \textbf{FEATS} & \textbf{HEAD} & \textbf{DEPREL} & \textbf{DEPS} & \textbf{NER} \\ \midrule

\multicolumn{10}{l}{ \# newdoc url = \url{http://www.poweredbyosteons.org/2012/01/brief-history-of-bioarchaeological.html}} \\
\multicolumn{10}{l}{\# newdoc s3 = \url{s3://aws-publicdatasets/common-crawl/crawl-data/CC-MAIN-2016-07/segments ...}} \\
\multicolumn{10}{l}{...} \\
\multicolumn{10}{l}{ \# sent\_id = \url{http://www.poweredbyosteons.org/2012/01/brief-history-of-bioarchaeological.html\#60}} \\
\multicolumn{10}{l}{\# text = The American Museum of Natural History was established in New York in 1869.} \\
0 & The & the & DT & DT & \_ & 2 & det & 2:det & O \\
1 & American & American & NNP & NNP & \_ & 2 & nn & 2:nn & B-Organization \\
2 & Museum & Museum & NNP & NNP & \_ & 7 & nsubjpass & 7:nsubjpass & I-Organization \\
3 & of & of & IN & IN & \_ & 2 & prep & \_ & I-Organization \\
4 & Natural & Natural & NNP & NNP & \_ & 5 & nn & 5:nn & I-Organization \\
5 & History & History & NNP & NNP & \_ & 3 & pobj & 2:prep\_of & I-Organization \\
6 & was & be & VBD & VBD & \_ & 7 & auxpass & 7:auxpass & O \\
7 & established & establish & VBN & VBN & \_ & 7 & ROOT & 7:ROOT & O \\
8 & in & in & IN & IN & \_ & 7 & prep & \_ & O \\
9 & New & New & NNP & NNP & \_ & 10 & nn & 10:nn & B-Location \\
10 & York & York & NNP & NNP & \_ & 8 & pobj & 7:prep\_in & I-Location \\
11 & in & in & IN & IN & \_ & 7 & prep & \_ & O \\
12 & 1869 & 1869 & CD & CD & \_ & 11 & pobj & 7:prep\_in & O \\
13 & . & . & . & . & \_ & 7 & punct & 7:punct & O \\ 
\multicolumn{10}{l}{...} \\

\bottomrule

\end{tabular}
\caption{An excerpt from an output document in the CoNLL format: a document header plus a sentence are shown. Here, ``ID'' is a word index, ``FORM'' is word form, ``LEMMA'' is lemma or stem of word form, ``UPOSTAG'' is universal part-of-speech tag, ``XPOSTAG'' is language-specific part-of-speech tag, ``FEATS'' is a list of morphological features, ``HEAD'' is head of the current word, which is either a value of ID or zero, ``DEPREL'' is universal dependency relation to the ``HEAD'', ``DEPS'' is enhanced dependency graph in the form of head-deprel pairs, and ``NER'' is named entity tag.
}
\label{tab:conll}
\end{table*}

\subsection{Linguistic Analysis of Texts}
Linguistic analysis consists of four stages presented in Figure~\ref{fig:outline} and is implemented using the Apache Hadoop framework\footnote{\url{https://hadoop.apache.org}} for parallelization and the Apache UIMA framework\footnote{\url{https://uima.apache.org}} for integration of linguistic analysers via the DKPro Core library~\cite{eckartdecastilho-gurevych:2014:OIAF4HLT}.\footnote{\url{https://github.com/uhh-lt/lefex}} 

\subsubsection{POS Tagging and Lemmatization}
For morphological analysis of texts, we used OpenNLP part-of-speech tagger and Stanford lemmatizer. 

\subsubsection{Named Entity Recognition}
To detect occurrences of persons, locations, and organizations we use the Stanford NER tool~\cite{finkel2005incorporating}.\footnote{stanfordnlp-model-ner-en-all.3class.distsim.crf, 20.04.2015} Overall, 7.48 billion occurrences of named entities were identified in the 251.92 billion tokens output corpus. 

\subsubsection{Dependency Parsing}
\label{sec:parsing}

To make large-scale parsing of texts possible, a parser needs to be not only reasonably accurate but also fast. Unfortunately, the most accurate parsers, such as Stanford parser based on the PCFG grammar~\cite{de2006generating}, according to our experiments, take up to 60 minutes to process 1 Mb of text on a single core, which was prohibitively slow for our use-case (details of the hardware configuration are available in Section~\ref{sec:hardware}). We tested all versions of the Stanford, Malt~\cite{hall2010single}, and Mate~\cite{ballesteros2014automatic} parsers for English available via the DKPro Core framework. To dependency-parse texts, we selected the Malt parser, due to an optimal ratio of efficiency and effectiveness (parsing of 1 Mb of text per core in 1--4 minutes). This parser was successfully used in the past for the construction of linguistically analyzed web corpora, such as \textsc{PukWaC}~\cite{baroni2009wacky} and \textsc{ENCOW16} ~\cite{Schaefer2015b}. While more accurate parsers exist, e.g. the Stanford parser, according to our experiments, even the neural-based version of this parser is substantially slower. On the other hand, as shown by~\newcite{chen2014fast}, the performance of the Malt parser is only about 1.5--2.5 points below the neural-based Stanford parser. In particular, we used the stack model based on the projective transition system with the Malt.\footnote{The used model is de.tudarmstadt.ukp.dkpro.core.maltparser-upstream-parser-en-linear, version 20120312.} 

The text downloaded from the Web has highly variable quality due to the inherent nature of user-generated content, but also unavoidable pre-processing errors, e.g. during the cleanup of incomplete HTML markup. To avoid crashes of the dependency parser caused by excessively long sentences, we filter all sentences longer than 50 tokens. Our manual analysis revealed that there are hardly any well-formed sentences of 50 tokens or more in this corpus. 

\subsubsection{Collapsing of Syntactic Dependencies}
Collapsed and enhanced dependencies, such as the Stanford Dependencies~\cite{de2006generating}\footnote{\url{https://nlp.stanford.edu/software/stanford-dependencies.shtml}} can be useful in various NLP tasks as they provide a more compact syntactic trees of a sentence, compared to the original dependency tree, thus reducing sparsity of syntax-aware representations.

To compensate the lack of the dependency enhancement in Malt, we use the system of~\cite{ruppert2015rule}\footnote{\url{http://jobimtext.org/dependency-collapsing}} to perform collapsing and enhancing of dependencies. The authors of the toolkit shown that (1) using the collapsed dependency representations substantially improves quality of construction of distributional thesauri based on sparse syntactic features; (2) the performance of the Stanford enhanced dependencies and the collapsed Malt dependencies on the same task are comparable. The advantage of using Malt with an external collapsing with respect to the Stanford parser, in our case, is speed. 

Note that, both original and enhanced versions are saved respectively into the columns ``DEPREL'' and ``DEPS'' as illustrated in Table~\ref{tab:conll}.

\subsection{Format of the Output Documents}
\label{sec:format}

The documents are encoded in the CoNLL format as illustrated  in Table~\ref{tab:conll}. The corpus is released as a collection of 19,101 gzip-compressed files. 

Each file is relatively small (around 150Mb) and is easy to download and work with locally during the development phase. However, to work with the entire corpus we recommend using some kind of parallelism, e.g. based on multiprocessing/multithreading or frameworks for distributed computing, such as Apache Hadoop/Spark/Flink.

\begin{figure*}[t]
\centering
\includegraphics[width=\textwidth]{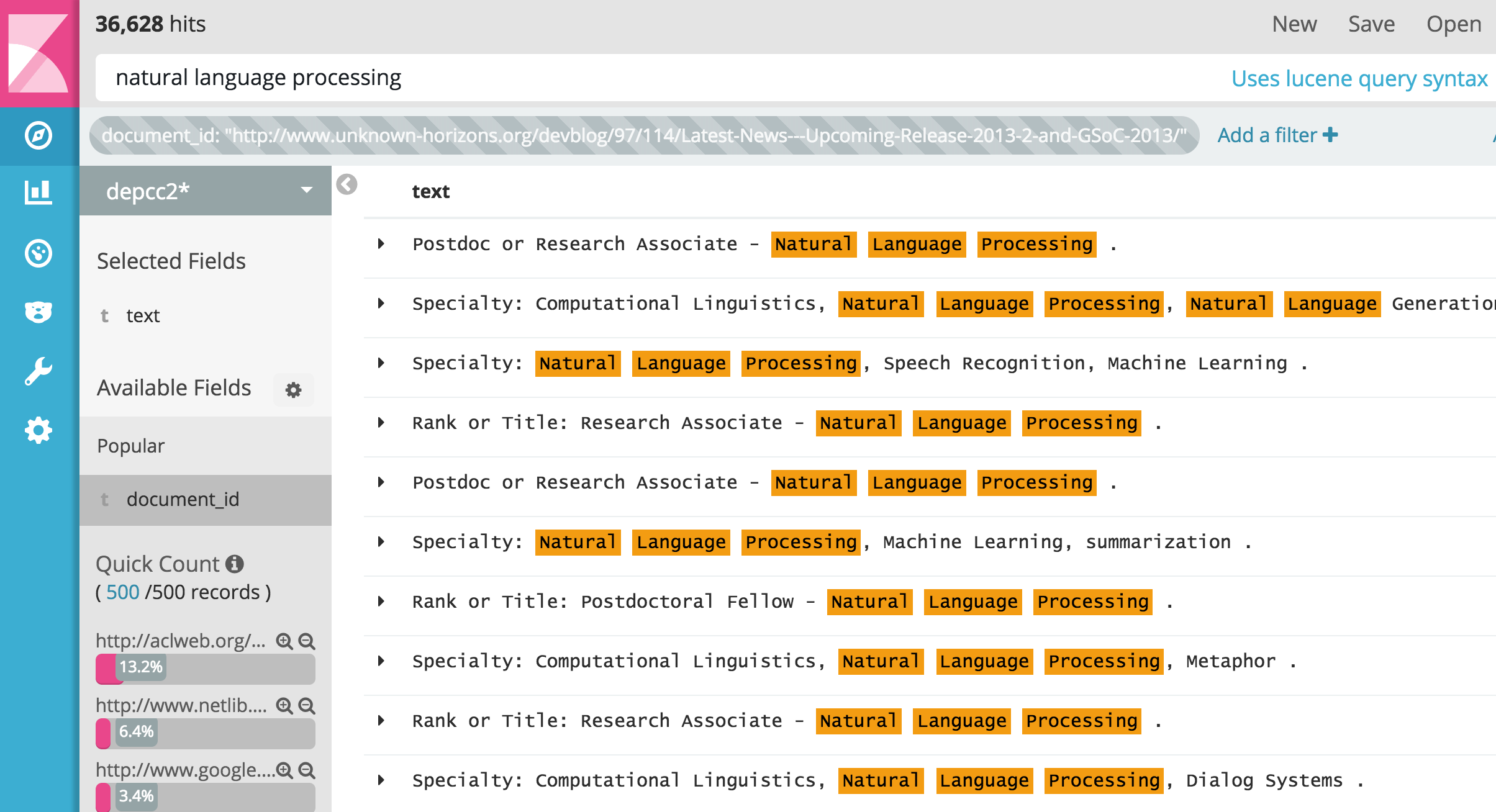}
\caption{Interactive graphical user interface providing a full text search over 14.3 billion of sentences in the \textsc{DepCC} corpus and their linguistic meta-data. A user can search for sentences containing specified keywords, named entities or syntactic dependencies.}
\label{fig:kibana}
\end{figure*}

\subsection{Computational Settings}
\label{sec:hardware}
The linguistic analysis was performed on an Apache Hadoop 2.6 cluster  using 341 containers each provided with one Intel Xeon CPU E5-2603v4@1.70GHz and 8Gb of RAM. The computational cluster consisted of 16 nodes plus a single head node. The job used 2.75 TB out of 2.82 TB available RAM and 356 out of 640 available Vcores. 

\subsection{Running Time}
In total, the computations were completed in 110 hours in 19101 tasks each processing a block of 100 Mb input data. The median running time of one task was 1 hour 10 minutes. This corresponds to the processing time of 
about 1.4 Mb/min for such a median task and 0.84 Mb/min for on average for the entire corpus, including compression of the output CoNLL files (cf. Section~\ref{sec:parsing}). The minimum time of processing  of a task was 38 minutes while the maximum time was 9 hours and 4 minutes.

\subsection{Using the Corpus in the Amazon Cloud}
\label{sec:amazon}

The \textsc{Common Crawl} datasets are hosted in the Amazon computing cloud platform.\footnote{\url{https://aws.amazon.com}} As mentioned in the introduction, our corpus is also was made directly available on Amazon S3 distributed file system in the \texttt{us-east-1} region (US East, North Virginia) as a part of the \textsc{Common Crawl} contributed datasets. This means that \textit{you do not need to download the corpus} to be able to work with it. Instead, you can run the jobs directly against the respective bucket on the Amazon S3 file system which contains the \textsc{DepCC} corpus. For optimal performance, you need to run instances which perform computations with the corpus inside the Amazon cloud (e.g. using the EC2\footnote{\url{https://aws.amazon.com/ec2}} or EMR\footnote{\url{https://aws.amazon.com/emr}} services) in the \texttt{us-east-1} region. 

\subsection{Index of the Corpus}

An access to a full-text search of all 14.3 billion sentences and their dependency relations of the \textsc{DepCC} corpus is  available upon request. This service is free and aims at facilitating access to the corpus for use-cases that do not require download of the entire collection of documents. Using the index, users can quickly retrieve sentences matching various linguistic criteria (using the Lucene query syntax\footnote{\url{https://www.elastic.co/guide/en/kibana/current/lucene-query.html}}), e.g. presence of keyword in a sentence, presence of a specific named entity in a sentence, presence of a specific dependency relation, provenance of a document from a specific web domain, etc.  Each retrieved sentence contains all the meta-data depicted in Table~\ref{tab:conll}, such as the provenance of the sentence.

The corpus can be queried via a RESTful API based on the ElasticSearch search engine\footnote{\url{https://www.elastic.co}} or via a web-based graphical user interface based on Kibana\footnote{\url{https://www.elastic.co/products/kibana}} graphical interface to ElasticSearch. Results of a sample query visualized using  Kibana  are presented in Figure~\ref{fig:kibana}.

We do not distribute the index itself due to its huge size. However, users can re-create the  index using the open source software provided as a part of the JoBimText package\footnote{\url{https://github.com/uhh-lt/josimtext}} from the the CoNLL files. While these compressed CoNLL files occupy only around 2.6 Tb, the size of the full index is about 15 Tb or more, depending on the replication factor of ElasticSearch. For this reason, for most practical applications,  re-creation of the index is faster and more straightforward than download of a pre-computed index and its subsequent deployment. Besides, some major versions of the ElasticSeach indices are not compatible between one another.

\begin{table*}[ht]
\footnotesize
\centering
\begin{tabular}{lcccc}
\toprule

\textbf{Model} &  \textbf{SimVerb3500} & \textbf{SimVerb3000} &  \textbf{SimVerb500} & \textbf{SimLex222}    \\ \midrule

Wikipedia+ukWaC+BNC: Count SVD 500-dim~\cite{baroni2014don} & 0.196 & 0.186 & 0.259 & 0.200 \\
PolyglotWikipedia: SGNS BOW 300-dim~\cite{gerz-EtAl:2016:EMNLP2016} & 0.274 & 0.333 & 0.265 & 0.328 \\
8B: SGNS BOW 500-dim~\cite{gerz-EtAl:2016:EMNLP2016} & 0.348 & 0.350 & 0.378 & 0.307  \\
8B: SGNS DEPS  500-dim~\cite{gerz-EtAl:2016:EMNLP2016} & \textbf{0.356} & \textbf{0.351} & 0.389 & 0.385 \\
PolyglotWikipedia:SGNS DEPS 300-dim~\cite{gerz-EtAl:2016:EMNLP2016} & 0.313 & 0.304 & \textbf{0.401} & \underline{\textbf{0.390}} \\
\midrule

Wikipedia: LMI DEPS wpf-1000 fpw-2000 & 0.283 & 0.284 & 0.271 & 0.268 \\
Wikipedia+ukWac+GigaWord: LMI DEPS wpf-1000 fpw-2000 & 0.376 & 0.368 & 0.419 & 0.183 \\ 

\textsc{DepCC}: LMI DEPS wpf-1000 fpw-1000 & 0.400 & 0.387 & \underline{\textbf{0.477}} & 0.285 \\ 
\textsc{DepCC}: LMI DEPS wpf-1000 fpw-2000 & \underline{\textbf{0.404}} & \underline{\textbf{0.392}} & \underline{\textbf{0.477}} & \textbf{0.292} \\ 
\textsc{DepCC}: LMI DEPS wpf-2000 fpw-2000 & 0.399 & 0.388 & 0.459  & 0.268 \\ 
\textsc{DepCC}: LMI DEPS wpf-5000 fpw-5000 & 0.382 & 0.372 & 0.442  & 0.226 \\ 
\bottomrule
\end{tabular}
\caption{Evaluation results on the verb semantic similarity task. Sparse count-based  distributional models (LMI) trained on the \textsc{DepCC} corpus are compared to models trained on the smaller corpora, such as Wikipedia and a combination of Wikipedia, \textsc{ukWac}, and \textsc{GigaWord}. Rows and columns of each LMI-weighted distributional model are pruned:
the \textit{wpf} indicates the number of words per feature, and the \textit{fpw} indicates the number of features per word. We also compare our models to the best verb similarity models from the state of the art. Here the ``BOW'' denotes models based on bag-of-word features, while ``DEPS'' denotes syntax-based models. SimVerb3000 and SimVerb500 are train and test partitions of the SimVerb3500, while the SimLex222 dataset is composed of verb pairs from the SimLex999 dataset.  The best results in a section are boldfaced, the best results overall are underlined. }
\label{tab:results}
\end{table*}

\section{Terms of Use}
\label{sec:license}
The \textsc{DepCC} corpus is based on a \textsc{Common Crawl} dataset. We do not reserve any copyrights as the authors of this derivative resource, but while using the \textsc{DepCC} corpus you need to make sure to respect the Terms of Use of the original \textsc{Common Crawl} dataset it is based on.\footnote{\url{http://commoncrawl.org/terms-of-use}}

\section{Evaluation: Verb Similarity Task}

As an example of potential use-case, we demonstrate the utility of the corpus and the overall methodology on a verb similarity task. 

This task structurally is the same as the word  similarity tasks based on such datasets as SimLex-999~\cite{hill2016simlex}. Namely, a system is given two words as input and needs to predict a scalar value which characterizes semantic similarity of the input words. While in the word similarity task the input pairs are words of various parts of speeches (nouns, adjectives, etc.), in this paper we only consider verb pairs. 

We chose this task since verb meaning is largely defined by the meaning of its arguments~\cite{fillmore1982frame}, therefore dependency-based features seem relevant for building distributional representations of verbs.

\subsection{Datasets: SimVerb3500 and SimLex222}
Recently a new challenging dataset for verb relatedness was introduced, called SimVerb3500~\cite{gerz-EtAl:2016:EMNLP2016}. The dataset is composed of 3500 pairs of verbs and is split into the train and test parts, called respectively SimVerb3000 and SimVerb500. In addition to this benchmark,  in our experiments, we also test the performance of the models on the SimLex222, which is the verb part of SimLex999 dataset~\cite{hill2016simlex} composed of 222 verb pairs. Historically, the SimVerb3500 dataset was created after the SimLex222, addressing its shortcomings related to the verb coverage. As in our experiments, we do not use the dataset SimVerb3000 for training, and to be consistent with the results reported in ~\cite{gerz-EtAl:2016:EMNLP2016} we report performance of the tested verb similarity models on all four datasets: SimVerb3500/3000/500, and SimLex222.

\subsection{A  Distributional Model for Verb Similarity}
\label{sec:jobimtext}
We compute syntactic count-based distributional representations of words using the JoBimText framework~\cite{Biemann:13}.\footnote{\url{https://github.com/uhh-lt/josimtext}} The sparse vectors are weighted using the LMI weighting schema and converted to unit length. In our experiments, we varied also the maximum number of salient features per word (\textit{fpw}) and words per feature (\textit{wpf}). Conceptually, each row and column of the sparse term-feature matrix is pruned such that at most \textit{wpf} non-zero elements  in a row and \textit{fpw} elements in a column are retained.

\subsection{Discussion of Results}

Table~\ref{tab:results} presents results of the experiments.

\subsubsection{Baselines}
The top part of the table lists five top systems in various categories~\cite{gerz-EtAl:2016:EMNLP2016}, representing the current state-of-art result on this dataset. Namely, the Count based SVD system is from ~\cite{baroni2014don}. In the original paper, two corpora were used: the ``8B'' is a 8 billion tokens corpus produced by a script in the word2vec toolkit, which gathers the texts from various sources~\cite{Mikolov:13} and the ``PolyglotWikipedia'' is the English Polyglot Wikipedia corpus consisting of 1.9 billion tokens~\cite{alrfou-perozzi-skiena:2013:CoNLL-2013}.

We use the baselines in the top of the table to indicate the best results on the dataset: our goal is to show the impact of the large corpora on performance and not to present a new model for verb similarity.

\subsubsection{Impact of the Corpora on Performance }
The bottom part of the table presents the distributional model described in Section~\ref{sec:jobimtext}~trained on the corpora of various sizes. Note, that the preprocessing steps for each corpus are exactly the same as for the \textsc{DepCC} corpus. We observe that the smallest corpus (Wikipedia) yields the worst results. While the scores go up on the larger corpus, which is a combination of Wikipedia with two other corpora, we can reach the even better result by training the model (with exactly the same parameters) on the dependency-based features extracted from the full \textsc{DepCC} corpus. This model substantially outperforms also the prior state of the art models, e.g. \cite{baroni2014don} and \cite{gerz-EtAl:2016:EMNLP2016}, on the SimVerb dataset, through the sheer size of the input corpus, as previously shown, e.g.~\cite{banko-brill:2001:ACL} inter alia. 

\subsubsection{Differences in Performance for Test/Train Sets}

For the SimVerb dataset, the absolute performance on the test part (SimVerb500) is higher than the absolute performance on the train part (SimVerb300) for almost all models, including the baselines. We attribute this to a specific split of the data in the dataset: our models do not use the training data to learn verb representations. 

\section{Conclusion}
\label{sec:conc}

In this paper, we introduced a new web-scale corpus of English texts extracted from the \textsc{Common Crawl}, the largest openly available linguistically analyzed corpus to date, according to the best of our knowledge. 

The documents were de-duplicated and linguistically processed with part-of-speech and named entity taggers and a dependency parser, making it possible to easily start large-scale experiments with syntax-aware models without the need of long and resource-intensive preprocessing. We built an index of sentences and their linguistic meta-data accessible though an interactive web-based search interface or via a RESTful API.

In our experiments on the verb similarity task, a distributional model trained on the new corpus  outperformed models trained on the smaller corpora, like Wikipedia, reaching new state of the art of verb similarity on the SimVerb3500 dataset. The corpus can be used in various contexts, ranging from training of syntax-based word embeddings~\cite{levy-goldberg:2014:P14-2} to unsupervised induction of  word senses~\cite{biemann2018framework} and frame structures~\cite{kawahara-peterson-palmer:2014:P14-1}. A promising direction of future work is using the proposed technology for building corpora in multiple languages.   

\section{Acknowledgements}

This research was supported by the Deutsche For\-schungs\-gemeinschaft (DFG) under the project "Joining Ontologies and Semantics Induced from Text" (JOIN-T). We are grateful to Amazon for providing required computational resources though the ``AWS Cloud Credits for Research'' program. Finally, we thank Kiril Gashteovski and three anonymous reviewers for their most helpful feedback.

\section{Bibliographical References}
\label{main:ref}

\bibliographystyle{lrec}
\bibliography{lrec}


\end{document}